\documentclass[sigconf]{acmart}

\usepackage{amsmath,amsfonts,bm}

\def\eqref#1{equation~\ref{#1}}

\def\1{\bm{1}}

\def\rvf{{\mathbf{f}}}

\def\rvl{{\mathbf{l}}}

\def\rvr{{\mathbf{r}}}

\def\rvx{{\mathbf{x}}}

\def\rmA{{\mathbf{A}}}

\def\rmO{{\mathbf{O}}}
\def\rmP{{\mathbf{P}}}
\def\rmQ{{\mathbf{Q}}}

\DeclareMathAlphabet{\mathsfit}{\encodingdefault}{\sfdefault}{m}{sl}
\SetMathAlphabet{\mathsfit}{bold}{\encodingdefault}{\sfdefault}{bx}{n}

\def\gA{{\mathcal{A}}}

\def\gC{{\mathcal{C}}}

\def\gM{{\mathcal{M}}}

\def\gS{{\mathcal{S}}}

\def\sR{{\mathbb{R}}}

\def\sZ{{\mathbb{Z}}}

\usepackage{graphicx}
    \graphicspath{{figs/}}
\usepackage{url}
\usepackage{multirow}
\usepackage[figuresright]{rotating}
\usepackage{amsmath}
\usepackage{colortbl}
\usepackage{color}
\usepackage{caption}
\usepackage{subcaption}
\usepackage{wrapfig}
\usepackage{makecell}
\usepackage{tabularx}
\usepackage[ruled,vlined,linesnumbered]{algorithm2e}
\SetKw{KwBy}{by}

\usepackage{xcolor}
\definecolor{MyBlue}{HTML}{11144C}
\definecolor{MyGreen}{HTML}{3A9679}
\definecolor{MyYellow}{HTML}{FABC60}
\definecolor{MyRed}{HTML}{E16262}
\definecolor{Hyper}{HTML}{B63618}
\definecolor{MyPurple}{HTML}{6F69AC}

\AtBeginDocument{%
  }

\setcopyright{acmlicensed}
\copyrightyear{2025}
\acmYear{2025}
\setcopyright{cc}
\setcctype{by-nc-sa}
\acmConference[MM '25]{Proceedings of the 33rd ACM International Conference
on Multimedia}{October 27--31, 2025}{Dublin, Ireland}
\acmBooktitle{Proceedings of the 33rd ACM International Conference on
Multimedia (MM '25), October 27--31, 2025, Dublin,
Ireland}
\acmISBN{979-8-4007-2035-2/2025/10}
\acmDOI{10.1145/3746027.3755370}

\begin{document}

\title{SizeGS: Size-aware Compression of 3D Gaussian Splatting via Mixed Integer Programming}

\author{Shuzhao Xie$^*$}
\affiliation{%
  \institution{SIGS, Tsinghua University}
  \city{Shenzhen}
  \country{China}
}
\email{xsz24@mails.tsinghua.edu.cn}

\author{Jiahang Liu$^*$}
\affiliation{%
  \institution{Harbin Institute of Technology}
  \city{Shenzhen}
  \country{China}
}
\email{liu030526@gmail.com}

\author{Weixiang Zhang}
\affiliation{%
  \institution{SIGS, Tsinghua University}
  \city{Shenzhen}
  \country{China}
}
\email{zwx.healthy@gmail.com}

\author{Shijia Ge}
\affiliation{%
  \institution{SIGS, Tsinghua University}
  \city{Shenzhen}
  \country{China}
}
\email{gsj23@mails.tsinghua.edu.cn}

\author{Sicheng Pan}
\affiliation{%
  \institution{SIGS, Tsinghua University}
  \city{Shenzhen}
  \country{China}
}
\email{scpan0306@gmail.com}

\author{Chen Tang}
\affiliation{%
  \institution{MMLab, The Chinese University of Hong Kong}
  \city{Hong Kong}
  \country{China}
}
\email{chentang@link.cuhk.edu.hk}

\author{Yunpeng Bai}
\affiliation{%
  \institution{The University of Texas at Austin}
  \city{Texas}
  \country{USA}
}
\email{byp215@gmail.com}

\author{Cong Zhang}
\affiliation{%
  \institution{Jiangxing Intelligence Inc.}
  \city{Shenzhen}
  \country{China}
}
\email{congz@ieee.org}

\author{Xiaoyi Fan}
\affiliation{%
  \institution{Jiangxing Intelligence Inc.}
  \city{Shenzhen}
  \country{China}
}
\email{xiaoyi.fan@ieee.org}

\author{Zhi Wang$^\dagger$}
\affiliation{%
  \institution{SIGS, Tsinghua University}
  \city{Shenzhen}
  \country{China}
}
\email{wangzhi@sz.tsinghua.edu.cn}
\thanks{$^*$Shuzhao Xie and Jiahang Liu contributed equally to this research.}
\thanks{$^\dagger$Corresponding author}

\renewcommand{\shortauthors}{Shuzhao Xie et al.}
\begin{abstract}
    Recent advances in 3D Gaussian Splatting (3DGS) have greatly improved 3D reconstruction. However, its substantial data size poses a significant challenge for transmission and storage. While many compression techniques have been proposed, they fail to efficiently adapt to fluctuating network bandwidth, leading to resource wastage. We address this issue from the perspective of size-aware compression, where we aim to compress 3DGS to a desired size by quickly searching for suitable hyperparameters. Through a measurement study, we identify key hyperparameters that affect the size—namely, the reserve ratio of Gaussians and bit-width settings for Gaussian attributes. Then, we formulate this hyperparameter optimization problem as a mixed-integer nonlinear programming (MINLP) problem, with the goal of maximizing visual quality while respecting the size budget constraint. To solve the MINLP, we decouple this problem into two parts: discretely sampling the reserve ratio and determining the bit-width settings using integer linear programming (ILP). To solve the ILP more quickly and accurately, we design a quality loss estimator and a calibrated size estimator, as well as implement a CUDA kernel. 
    Extensive experiments on multiple 3DGS variants demonstrate that our method achieves state-of-the-art performance in post-training compression. Furthermore, our method can achieve comparable quality to leading training-required methods after fine-tuning. Project page \& Code: \href{https://shuzhaoxie.github.io/sizegs}{shuzhaoxie.github.io/sizegs}.
\end{abstract}



\maketitle

\section{Introduction}
\label{sec:intro}

In recent years, 3D Gaussian Splatting (3DGS)~\cite{kerbl20233d} has revolutionized 3D scene representation and has been widely adopted in a variety of applications~\cite{wang2024v3,li2024reality,chen2025lara,ziwen2024long,li2024nerfcodec,li2024tuning,gu2024drag,yao2025sdgs,tan2024watergs}. Despite its success, 3DGS still faces limitations in transmission efficiency due to its giant number of points and complex attributes. Though various 3DGS compression methods have been proposed~\cite{3DGSzip2024}, they overlook requirements arising from on-demand applications such as volumetric video streaming~\cite{wang2024v3,zhang2022neuvv}, live streaming~\cite{10.1145/3570361.3592530}, and remote teleoperation~\cite{li2024reality}. These applications frequently encounter fluctuating network bandwidth, leading to jitter and blurriness that significantly impact user experience. These inefficiencies are further exacerbated in large-scale 3DGS scenes, which often exceed 20 megabytes (MB) even after compression. 
To overcome this challenge, size-aware 3DGS compression presents a promising direction. The key idea is to \emph{compress the 3DGS to desired size by automatically searching the suitable hyperparameter set}.

Based on the way of hyperparameter configuration, current 3DGS compression methods fall into two categories. One is the \emph{offline} method~\citep{morgenstern2023compact,papantonakis2024i3d,wu2024implicitgs,yang2024sundae,xie2024mesongs,niedermayr2023compressed,fan2023lightgaussian}, which requires manually setting hyperparameters to adjust the compressed file size. Moreover, there are a variety of hyperparameters to consider, which demands a significant amount of human effort for adjustment. The other is the \emph{online} method~\citep{lee2023compact,liu2024compgs,hac2024,wang2024contextgs,wang2024rdo}, which selects hyperparameters based on a context model and imports an extra hyperparameter $\lambda$ to balance rate and bit consumption. While $\lambda$ can adjust the final size, each adjustment requires retraining the context model from scratch, which is time-consuming. Additionally, it is difficult to predict the final size given a certain $\lambda$. Therefore, existing methods are unable to search for feasible hyperparameters under the constraint of the size budget in a short amount of time. Given the fast speed of recent integer programming solvers~\cite{yao2021hawq}, we propose to achieve fast size-aware compression via mixed integer programming.

However, searching for appropriate 3DGS compression hyperparameters from an optimization perspective is non-trivial. First, 3DGS compression pipeline has lots hyperparameters. Some hyperparameters have a significant impact on size, while others are not. It is necessary to filter out a set of hyperparameters that significantly influence size. Second, to formulate an integer programming model, an explicit analytical relationship needs to be established between size and hyperparameters, as well as between quality and hyperparameters. This model can then use existing solvers to determine the appropriate hyperparameter settings. Finally, to fastly achieve the accurate results, an accurate approximation function for the actual size and acceleration techniques should be propose, which will allow the solver determine whether a candidate hyperparameter solution satisfies the constraints faster.

To tackle these challenges, we propose \emph{SizeGS}, a codec that can compression 3DGS to a desired file size while maximizing visual quality. First, we summarize and analyze the pipeline of 3DGS compression and identify the two most influential hyperparameters on size: the \textbf{reserve ratio $\tau$} and \textbf{bit-width setting $\rmQ$}. Then, we formulate the problem as a Mixed Integer Nonlinear Programming (MINLP) model, where the objective is visual quality and the constraint is size. To solve it, we decompose the MINLP into two subproblems: one is discrete sampling for $\tau$, and the other is Integer Linear Programming (ILP) for $\rmQ$, which can be quickly solved using existing solvers~\cite{pulp}. Specifically, we perform multiple rounds of search to find suitable hyperparameters. In each round, we sample a value for $\tau$, solve for $\rmQ$ using ILP, compute the quality. We select the hyperparameters with the  highest quality as the search result. To ensure that the bit-width settings search aligns with the ILP definition (i.e., both objective and constraint are linear functions of $\rmQ$), we propose using quantization loss to replace visual quality and construct a linear analytical relationship between size and $\rmQ$ for accurate size estimation. To accelerate the solution process, we implement a CUDA kernel for parallel quantization and reuse the search results from the previous round as the solver’s search starting point. Finally, we carefully design a three-step fine-tuning method to improve accuracy. 

In summary, our contributions are as follows: \textbf{1)} We formulate the hyperparameter search problem in 3DGS compression as a Mixed Integer Nonlinear Programming (MINLP) model. To make this optimization tractable, we decouple this problem into two subproblems: discrete sampling for the reserve ratio and Integer Linear Programming (ILP) for bit-width setting. \textbf{2)} To enable precise size estimation, we establish an analytical linear relationship between the bit-width setting and the resulting file size. To accelerate the ILP solver, we utilize the attribute loss to estimate the quality loss and implement a CUDA kernel to parallelize the quantization process.
\textbf{3)} Extensive experiments on multiple 3DGS variants reveal that our method can search a set of hyperparameters to compress 3D Gaussians to desired size in a minute and achieves SOTA performance on offline compression. With finetuning, our method can achieve the comparable performance as the SOTA online compression methods.
 
\section{Preliminary and Motivation}

In this section, we first introduce the the data composition of 3DGS. Then, we summarize the general techniques of online and offline 3DGS compression methods. Finally, we identify the hyperparameters that greatly influence the compressed file size.

\subsection{Base Models for 3DGS Compression}

\vspace{1mm}\noindent\textbf{3DGS} \citep{kerbl20233d} consists of mulitple 3D Gaussian functions. Each Gaussian is characterized by a covariance matrix $\mathbf{\Sigma}$ and a center point $\mu$, which is referred to as the mean value of the Gaussian: $G(x) = e^{-\frac{1}{2}(x-\mu)^\top \mathbf{\Sigma}^{-1}(x-\mu)}$.
To maintain the positive definiteness of the covariance matrix $\mathbf{\Sigma}$, 
3DGS decomposes $\mathbf{\Sigma}$ into a scaling matrix $\mathbf{S} = {\rm diag}(\mathbf{s}), \mathbf{s} \in \mathbb{R}^3$ and a rotation matrix $\mathbf{R}$: $\mathbf{\Sigma} = \mathbf{R}\mathbf{S}\mathbf{S}^\top \mathbf{R}^\top$.
The rotation matrix $\mathbf{R}$ is parameterized by a rotation quaternion $\mathbf{q} \in \mathbb{R}^4$. In summary, each element of 3D Gaussians constains:
(1) a 3D center $\mathbf{\mu} \in \mathbb{R}^3$; (2) a rotation quaternion $\mathbf{q} \in \mathbb{R}^4$;
(3) a scale vector $\mathbf{s} \in \mathbb{R}^3$; 
(4) a color feature defined by spherical harmonics (SH) coefficients $\mathbf{SH} \in \mathbb{R}^h$, with $h = 3(d+1)^2$, where $d$ is the harmonics degree; 
and (5) an opacity logit $o \in \mathbb{R}$. 

\vspace{1mm}\noindent \textbf{Scaffold-GS}~\cite{lu2024scaffold} is a variant of 3DGS, widely adopted in 3DGS compression~\cite{hac2024,wang2024contextgs}. It introduces \textit{anchor points} to capture common attributes of local 3D Gaussians. Specifically, the \textit{anchor points} are initialized from neural Gaussians by voxelizing the 3D scenes. Each anchor point has a context feature $\rvf \in \sR^{32}$, a location $\rvx \in \sR^3$, a scaling factor $\rvl \in \sR^6$ and $k$ learnable offset $\rmO \in \sR^{k\times 3}$. 

\vspace{1mm}\noindent \textbf{4DGS}~\cite{yang2023gs4d} models dynamic scenes as spatio-temporal 4D volumes composed of 4D Gaussian primitives. Each 4D Gaussian consists of (1) a 3D center $\mathbf{\mu} \in \mathbb{R}^3$; (2) a rotation quaternion $\mathbf{q} \in \mathbb{R}^4$;
(3) a scale vector $\mathbf{s} \in \mathbb{R}^3$; 
(4) a color feature defined by SH coefficients $\mathbf{SH} \in \mathbb{R}^h$, with $h = 3(d+1)^2$, where
$d$ is the harmonics degree; (5) an opacity logit $o \in \mathbb{R}$; (6) a time coordinate $\mathbf{t} \in \sR$; (7) a time scale component $\mathbf{s}_t \in \mathbb{R}$; (8) a time-level rotation vector $\mathbf{r}_t \in \mathbb{R}^4$.

\vspace{1mm}\noindent \textbf{Summary.} Elements of base 3D Gaussian models can be divided into two components: \emph{geometry} and \emph{attributes}. The \emph{geometry} consists of the coordinates of 3DGS. The \emph{attributes} consists of the other components. For example, for ScaffoldGS, the \emph{geometry} refers to the coordinates of the anchor points, and the \emph{attributes} refers to the context feature, scaling factors and learnable offsets.

\begin{figure}[t]
  \centering
  \begin{subfigure}{0.49\linewidth}
      \includegraphics[width=0.99\linewidth]{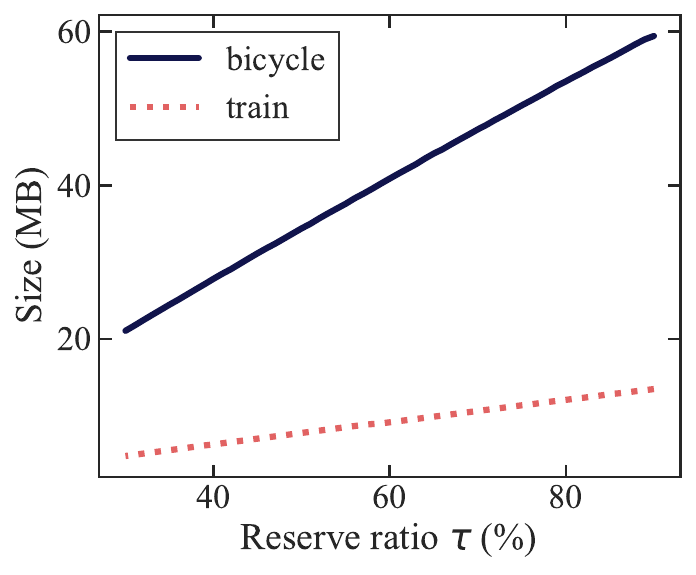}
    \caption{Size vs. Reserve ratio.}
    \label{fig:size_prune}
  \end{subfigure}
  \hfill
  \begin{subfigure}{0.49\linewidth}
      \includegraphics[width=0.99\linewidth]{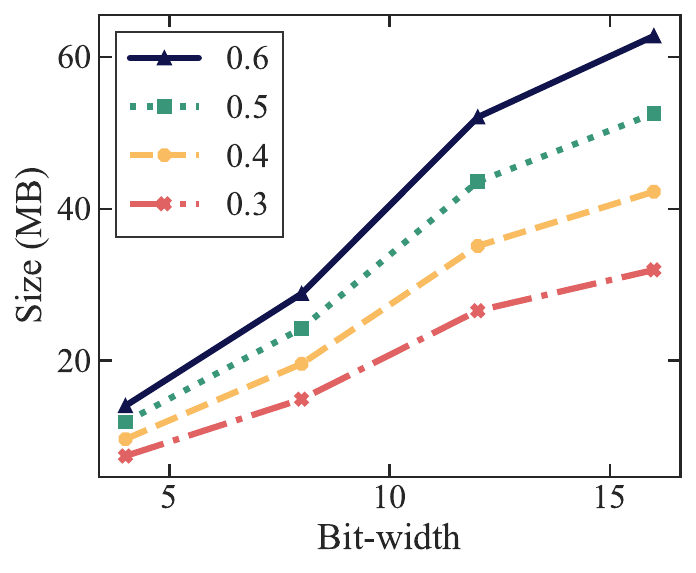}
    \caption{Size vs. Bit-width.}
    \label{fig:size_bit}
  \end{subfigure}
  \vspace{-0.4cm}
  \caption{The reserve ratio and bit-width strongly impact file size and exhibit a near-linear relationship with it.}
  \label{fig:op_mix}
  \vspace{-0.5cm}
\end{figure}

\subsection{Universal Pipeline of 3DGS Compression}\label{sec:tech}

3DGS compression pipeline consists of four steps: 1) pruning the unimportant Gaussian points, 2) Transforming the attributes (i.e., quaternions) to reduce data entropy; 3) Quantizing the attributes in a group-wise way; 4) Entropy coding the coordinates and attributes.

\vspace{1mm}\noindent \textbf{Pruning.} Offline methods first prune the unimportant Gaussians based on a importance score~\cite{fan2023lightgaussian,niedermayr2023compressed,xie2024mesongs}, derived from the volume rendering function: $C = \sum_{i\in N}{c_i \alpha_i \prod_{j=1}^{i-1}(1 - \alpha_j)}$. Here, $c_i$ and $\alpha_i$ are the density and color of this pixel, computed by a Gaussian with covariance $\mathbf{\Sigma}$ multiplied by
a per-point opacity and SH coefficients. Based on this function, the importance score is given by:
\vspace{-0.3cm}
\begin{equation}
  \label{eq:imp_meson}
  \quad I_d = \sum_{p \in \mathcal{P}} \alpha_i \prod_{j=1}^{i-1}(1 - \alpha_j),
  \vspace{-0.2cm}
\end{equation}
$\mathcal{P}$ is the pixel set that is overlapped by the Gaussian $g$,
and $i$ is the rank of Gaussian $g$ in a set of Gaussians that overlap with the pixel $p$. It uses a threshold $\tau$ to prune Gaussians, which means that it retains the percent of $\tau$ of the sorted Gaussians. Some works~\cite{fan2023lightgaussian,xie2024mesongs} use volume to weight $I_d$ to obtain a more precise importance estimation, but this improvement is negligible. Here the volume means the product of the scale vector. 

\vspace{1mm}\noindent \textbf{Transformation.} This technique is usually employed by offline compression method. A typical method is region-adaptive hierarchical transform (RAHT)~\cite{de2016compression}, which is utilized by MesonGS~\cite{xie2024mesongs} to decompose \emph{attributes} into low-frequency and high-frequency coefficients. This step can reduce the entropy of the attributes, resulting in less information loss during subsequent quantization and leading to better entropy encoding results. 

\vspace{1mm}\noindent \textbf{Quantization.} Due to the extreme sensitivity of geometry, previous works typically quantize it to 16-bit. For attributes, previous methods partition each channel of attributes into multiple groups. For each group of attributes, there are currently two types of quantization approaches. The first is based on traditional compression models, where the quantization step $Q_s \in [0, 1]$ are used to convert the float attributes into integers: $\hat{x} = \lfloor x / Q_s \rceil$. The second approach adopts model-based quantization, which quantize a group of attributes $x$ with the quantization bit-width $Q_b \in [1, 32] \cap \sZ$:
\begin{equation}
  \hat{x} = \lfloor {\rm clamp}(x / s_x + z_x, 0, 2^{Q_b} - 1) \rceil,
\end{equation}
where $s_x = [\max(x) - \min(x)] / 2^{Q_b}$ and $z_x = \lfloor 2^{Q_b} - \max(x) / s_x \rceil$.
Here, $\lfloor \cdot \rceil$ represents the rounding-to-nearest function, and $\hat{x}$ refers to the quantized attributes. Besides, function ${\rm clamp}(\cdot)$ specifies a range of values. Values below the minimum are set to the minimum. Values above the maximum are set to the maximum. Without specification, we use the notation $\rmQ$ to denote the bit-width settings and $Q$ to denote the number of possible bit-widths.

\vspace{1mm}\noindent \textbf{Entropy Coding.} For quantized geometry, most existing methods utilize G-PCC~\cite{pcc_overview} for entropy coding. For quantized attributes, some approaches~\cite{fan2023lightgaussian,xie2024mesongs,niedermayr2023compressed} directly compress them using LZ77 codecs. Other methods~\cite{hac++2025,liu2024compgs} model the probability distributions of quantized attributes as Gaussian Mixture Model (GMM) to provide additional information for entropy coding and thereby achieve higher compression rates. Specifically, they estimate the mean and variance for each quantization group to construct the GMM.

\subsection{Motivation}

\textbf{Size-sensitive Hyperparameters.} Among the above four steps, three of them require configuring the hyperparameters. These hyperparameters includes the reserve ratio $\tau$, bit-width setting $\rmQ$, the number of quantization groups $B$, and the mean and variance used for constructing the probability distribution of entropy coding. Among these hyperparameters, as depicted in Fig.~\ref{fig:op_mix}, the reserve ratio $\tau$ and the bit-width setting $\rmQ$ show great impact on the final file size. The number of quantization groups $B$ and the bit-width setting $\rmQ$ are actually inversely related, so only one needs to be retained for the search. Entropy encoding is used as the final step of the entire pipeline, and its corresponding hyperparameters have minimal impact on the final size~\cite{hac++2025,zhan2025catdgs}. Therefore, we do not consider the hyperparameters of entropy encoding or the number of blocks during the hyperparameters optimization.

For \emph{online compression}, hyperparameters are configured via a learned mask and context model: $\tau$ from a learned mask, $\rmQ$ and distribution parameters from a context model. The online method balances rate and quality via $\lambda$, but adjusting hyperparameters requires retraining the context model. Hence, searching by size is time-consuming. For \emph{offline compression}, all these hyper-parameters need to be manually configured. Inspired by linear programming solvers~\cite{pulp,fastdog2022}, we frame hyperparameter search as a planning problem and introduce techniques to speed up verification, enabling rapid iterative search.

\section{Methodology}
\begin{figure*}[t]
    \centering
     \includegraphics[width=\linewidth]{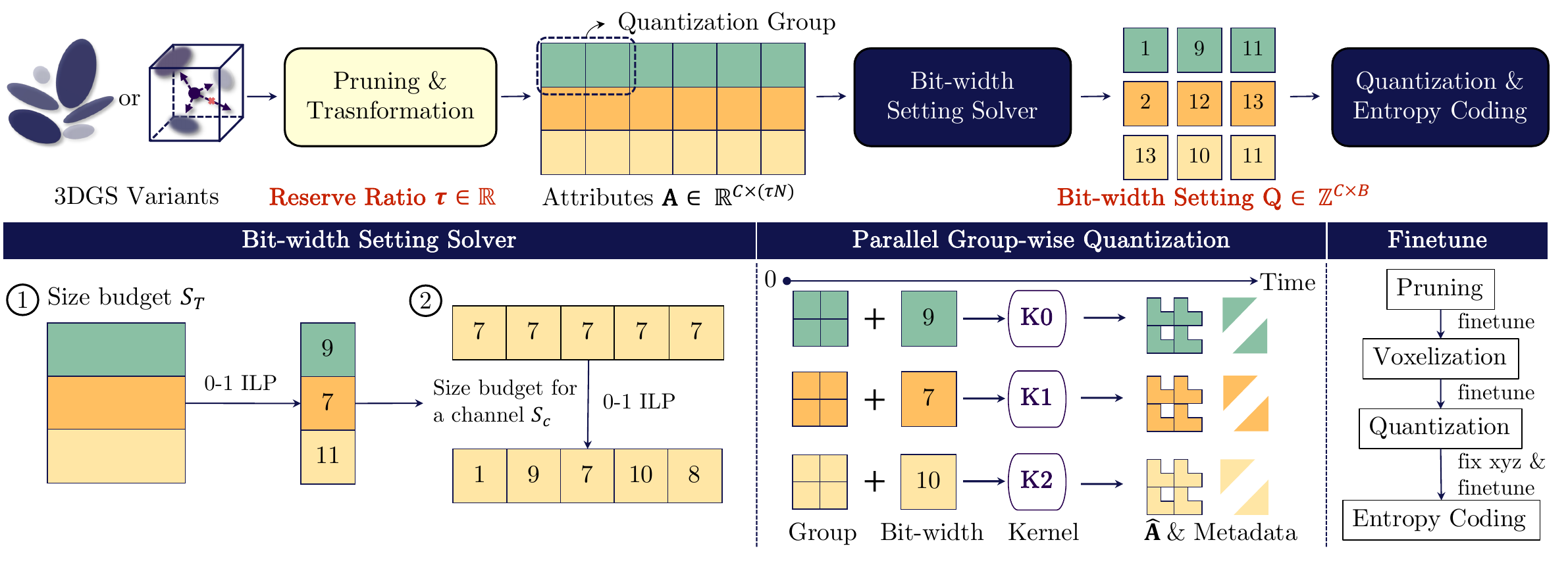}
     \vspace{-0.8cm}
     \caption{\textbf{Size-aware compression pipeline.} We achieve size-aware 3DGS compression via hyperparameter search. As illustrated above, the compression process can be controlled by adjusting the \textcolor{Hyper}{reserve ratio $\tau$} and \textcolor{Hyper}{bit-width setting $\rmQ$}. Algo.~\ref{algo:searching} presents our hyperparameter optimization algorithm, which aims to meet the size target while maximizing quality. To accelerate the bit-width setting solver, we use 0-1 ILP to compute the bit-width for each channel, and then construct a 0-1 ILP per channel to solve for the bit-width of each group. Additionally, we implement parallel group-wise quantization to further speed up the process. Finally, we restore quality via piecewise fine-tuning after pruning, voxelization, and quantization.}
    \label{fig:overview}
    \vspace{-0.3cm}
\end{figure*}

\textbf{Overview.} We present the compression pipeline and formulate size-aware 3DGS compression as a MINLP to find optimal reserve ratio $\tau$ and bit-width $\rmQ$. We split it into two parts: (1) discrete $\tau$ sampling, and (2) solving $\rmQ$ via ILP with fixed $\tau$. We derive an analytical $\rmQ$-to-size mapping, model it as a 0-1 ILP for accuracy, and refine size estimates iteratively. To speed up ILP, we use quantization loss to estimate quality drop and accelerate group-wise quantization via CUDA. Finally, we describe the finetuning process.

\subsection{Problem Formulation and Decoupling} \label{sec:iss}

\vspace{1mm}\noindent\textbf{Compression pipeline.} As shown in Fig.~\ref{fig:overview}, given a pre-traind 3DGS, we prune the Guassian points based on the importance score and percentage $\tau$, which means that $\tau N$ of the most important points are reserved. To calculate the importance of anchor points in ScaffoldGS, we average the importance of the corresponding generated Gaussian splats. Given the anchor point $a$, the set of training viewpoints $\mathcal{V}_a$ from which $a$ is visible, and the corresponding generated Gaussian points $\mathcal{G}_a$, the importance score $I_a$ is: $I_a = \sum_{v \in \mathcal{V}_a} \sum_{g \in \mathcal{G}_a} m_{g} I_d$, where $m_g \in \{1, 0\}$ reflects whether Gaussian $g$ is retained or discarded under viewpoint $v$. $I_d$ aggregates contributions across pixels overlapped by $g$, as shown in Eq.~\ref{eq:imp_meson}. For 3DGS and 4DGS, we use the Eq.~\ref{eq:imp_meson}.

We then quantize geometry and attributes. Following previous works~\cite{hac2024,liu2024compgs,hac++2025}, coordinates are set to 16-bit and compressed using G-PCC~\cite{pcc_overview}.Attribute tensor $\rmA \in \mathbb{R}^{ C \times (\tau N) }$ is quantized group-wise by splitting the first dimension into $C$ parts and the second into $B$ nearly equal parts, forming $C \times B$ \textbf{quantization groups}. We adopt the bit-width to quantize the attribute. Compared to the quantization step, a continuous variable, the bit-width is discrete, resulting in a smaller search space that is easier for optimization. Quantized attributes are then compressed with LZ77 or torchac~\cite{mentzer2019practical}. The final file includes geometry, attributes, and metadata (voxel size, block count, and per-group scale/zero points). 

\vspace{1mm}\noindent\textbf{Problem Formulation.} Given the above pipeline, our goal is to search for suitable values of the $\tau$ and $\rmQ$ under a given size constraint, while maximizing visual quality. Hence, we formulate this problem as a mixed integer nonlinear programming (MINLP) model:
\begin{equation}
\begin{aligned}
\label{eq:nilp}
& \underset{\mathbf{\tau}, \rmQ}{\text{minimize}} 
& & \gM(\tau, \rmQ),\\
& \text{subject to} 
& & \gS(\tau, \rmQ) \leq \text{Size Budget}, \\
&&& \tau \in [0, 1], \\
&&&  \rmQ \in [1, 32]^{C\times B} \cap \sZ^{C\times B}.
\end{aligned}
\end{equation}
Here, $\gM(\cdot)$ and $\gS(\cdot)$ are the quality loss and the estimated size of compressed model under the configuration of $\{\tau, \rmQ\}$, respectively. ``Mixed'' in MINLP means that the variable types include both discrete ($\rmQ$) and continuous ($\tau$) variables . A detailed introduction of MINLP are proposed in the supplementary material.

\vspace{1mm}\noindent\textbf{Structure Decoupling.} Since both $\gM(\cdot)$ and $\gS(\cdot)$ are nonlinear functions of $\{\tau, \rmQ\}$, and the search space is extremely large, solving this problem directly is highly challenging. For example, even when $C=8$ and $B=8$, the discrete combinatorial space already contains $32^{64}$ possibilities—far beyond astronomical scale! Moreover, optimizing the continuous variable $\tau$ within this vast space further increases the difficulty. Besides, existing MINLP solvers~\cite{scip,baronsolver} often require hours to find a reasonable solution. 

However, we observe only $\tau$ is continuous among these variables, while others are integers. Besides, by fixing $\tau$, we find a linear relationship between $\gS(\cdot)$ and $\rmQ$. E.g., as shown in Fig.~\ref{fig:size_bit}, using 8-bit across all groups yields proportional size, and 16-bit files are roughly twice as large as 8-bit ones—allowing size estimation from lower-bit results. Hence, if we fix the value of $\tau$ and approximate $\gM$ and $\gS$ as functions that are linearizable with respect to $\rmQ$, then this problem can be decoupled into two problems: 1) using discrete sampling to search and fix the value of $\tau$, and then 2) solve the $\rmQ$ via integer linear programming (ILP). The advantage of such a decoupling lies in the efficiency of recent ILP solvers~\cite{pulp,fastdog2022}, which requires a few seconds to obtain a solution.

\begin{algorithm}[t]
    \small
    \SetKwFunction{ILPSearch}{\texttt{01-ILP}}
    \KwIn{Size budget $S_T$ and a pre-trained 3DGS model}
    \KwOut{Hyperparameter set $\Phi^{\ast} = \{\tau^{\ast}, \rmQ^{\ast}\}$}
    Initialize bit-width setting $\rmQ \leftarrow \{8\}^{C \times B}$\;
    Initialize the best quality $M^{\ast} \leftarrow 0$\;
    \For{$\tau \in  \{\tau_1, \tau_2, \tau_3, ...\}$}{
        Compress model with current hyperparameters, obtain size $S_a$\;
        \If{$2 \times S_a < S_T$}{
            {\bf continue}\;
        }
        \While{true}{
            $S_{\Delta} \leftarrow S_a - S_T$\;
            \tcc{\ILPSearch($\cdot$) solver takes $\rmQ$ and $S_{\Delta}$ as input.}
            Search bit-widths via 01-ILP: $\rmQ \leftarrow$ \ILPSearch{$\mathrm{Q}, S_{\Delta}$}\;
            Compress model with $\tau$ and updated $\rmQ$, obtain new size $S_a$\;
            \If{$\frac{|S_a - S_T|}{|S_T|} < 0.05$}{
                {\bf break}\;
            }
        }
        \tcc{Retain the one with the best quality.}
        \If{$\Omega(\rmQ) > M^{\ast}$}{
            $M^{\ast} \leftarrow \Omega(\rmQ)$\;
            $\Phi^{\ast} \leftarrow \{\tau, \rmQ\}$\;
        }
    }
    {\bf return} {$\Phi^{\ast}$}\;
    \caption{\bf Hyperparameter Optimization}
    \label{algo:searching}
\end{algorithm}

\subsection{Solve the MINLP}\label{sec:ilp}

\textbf{Discrete Sampling of $\tau$.} Based on the above observations, we propose to traverse the value of reserve ratio $\tau$. Then, we can search the bit-width settings $\rmQ$ by solving the ILP. Hence, the hyperparameter searching algorithm for size-aware 3D Gaussian compression are proposed in Algo.~\ref{algo:searching}. The input is a target size and a pretrained 3DGS, and the output is the optimal hyperparameters set $\Phi^{\ast} = \{\tau^{\ast}, \rmQ^{\ast}\}$ that can satisfy the size constraint while maximizing visual quality. We iterate over all possible values of $\tau$. Here, $2 \times S_a < S_T$ refers to the situation where, if doubling the bit-width setting under the current configuration (up to a maximum value of 16) cannot achieve the target size, it indicates that $\tau$ is too small, and thus, more Gaussian points need to be retained.
For the bit-width setting search, we formulate the problem as a 0-1 Integer Linear Programming (ILP) task. The ILP solver takes $\rmQ$ and $S_{\Delta}$ as inputs, where $\rmQ$ serves as the initialization for each search iteration, enabling faster convergence, and $S_{\Delta}$ is used to calibrate the estimated size for more accurate results. After each iteration, the newly determined bit-width setting is used to store the results. If the stored result size $S_a$ is sufficiently close to the target size $S_T$, the search is terminated. For each pair of $\{\tau, \rmQ\}$, we compute the estimated quality $\Omega$ and retain the pair that yields the highest quality. We will introduce $\Omega$ below.

\vspace{1mm}\noindent\textbf{Binary ILP for $\rmQ$.} The input of mixed precision quantization is the important attributes $\gA$, which can be seen as a 2D-matrix. We divide this 2D-matrix into $C \times B$ blocks and there are $Q$ quantization options for each attribute block (e.g., 16 options for 1--16 bits). The search space of the ILP problem is $(C \times B)^Q$. 
The objective of solving the ILP is to find the best bit configuration in this search space that optimally balances quality loss $\Omega$ and the a size limit $\gS$. Besides, we define the bit-width variables as $\rmQ \in \{0, 1\}^{C \times B \times Q}$, which means that we use a one-hot vector $v \in \{0, 1\}^{Q}, |v| = 1$ to represent the bit-width setting for an attribute block. In all, the 0-1 ILP model tries to find the right bit-width setting $\rmQ$ can be formulated as
\begin{equation}
\begin{aligned}
\label{eq:ilp}
    & \underset{\rmQ}{\text{minimize}} 
    & & \Omega(\rmQ),\\
    & \text{subject to} 
    & & \gS(\rmQ) \leq \text{Size Budget}, \\
    &&&  \forall (i, j) \in [0, C) \times [0, B),  \sum_{q=1}^{Q} \rmQ_{i,j,q} = 1.
\end{aligned}
\end{equation}
\vspace{-2pt}
Here, $\gS(\rmQ)$ and $\Omega(\rmQ)$ denote the estimated file size and estimated quality loss under a bit-width setting $\rmQ$, respectively. Next, we introduce the details of the $\Omega$ and $\gS$.

\vspace{1mm}\noindent\textbf{Acceleration.} We propose the following techniques to accelerate the problem solving. Since computing the quality metric like PSNR requires traversing the entire training set and evaluating the metric, it takes at least 10 seconds, which significantly slows down the search process. To solve this, we use quantization loss $\Omega(\cdot)$ to replace the metric function $\gM(\cdot)$. Here, we assume that the bit-width of each group are independent of one another. This allows us to precompute the quantization loss of each group of attributes separately, and it only requires $Q$ times quantizations. As for the metric of quality loss, we use the distance between the original attributes and the restored attributes\footnote{Similar assumption can be found in~\cite{dong2019hawq,dong2019hawqv2}.}. Formally, we can precompute the estimated quality loss matrix $\Omega \in \mathbb{R}^{C \times B \times Q}$ with:
\begin{equation}
    \Omega(i, j, b) = | \hat{\mathcal{A}}_{i,j}^b - \mathcal{A}_{i, j} |.
\end{equation}
The $|\cdot|$ can be 1-norm, 2-norm, or $\infty$-norm. We plot the relationship between PNSR and $\Omega$ in the supplementary material, which reveals that minimizing $\Omega$ is equal to maximizing PSNR.
Besides, we set $Q$ as 16 to prune the search space. Finally, we implement a CUDA kernel to accelerate the quantization process, in which each quantization group is quantized in parallel, as shown in Fig.~\ref{fig:overview}.
 
\label{sec:estimator}
\vspace{1mm}\noindent\textbf{Size Estimator.} A compresssed 3DGS file constains the following components: 1) voxelized coordinates, 2) quantized attributes, and 3) metadata, which is used to restore the coordinates and attributes. Fortunately, we can store the voxelized coordinates and metadata to obtain the accurate compressed size in a few seconds. Then the challenge of size estimator lies in estimating the size of quantized attributes. This is to say, we only have to establish a analytical relationship between the compressed file size and the bit-width settings, the actual size of other components can be obtained by saving them to storage. Besides, the size estimator must be a linear function of the bit-width variables.

According to information theory~\cite{Elements_of_information_theory}, the lower bound of bit consumption can be calculated by $ \tau N \times (-\sum_{i}p_i \log_2{p_i})$. However, such a size estimator is not suitable for the formulation of ILP. The reason is that, for each searching iteration, we have to quantize the attributes into integers with the candidate bit-width setting and calculate the probability of each values to derive the bit consumption, which costs a lot of time. Moreover, the relationship between the bit-width settings and the estimated size is non-linear, which cannot satisfy the linear requirement of the ILP. Hence, an explicit and linear relationship between the size $\gS$ and the bit-width setting $\rmQ$ must be established.
Thus, we estimate the size by
\vspace{-6pt}
\begin{equation}
    \begin{aligned}
        \label{eq:size_estim}
            \gS(\rmQ) = \sum_{i,j} \rmP_{ij} \rmQ_{ij} + \gC + S_\Delta.
    \end{aligned}
    \vspace{-12pt}
\end{equation} 
Here, $\rmP \in \mathbb{R}^{C \times B}$ refers to size of quantization groups. $\mathcal{C}$ refers to the accurate storage consumption of the metadata and the coordinates, which can be obtained by storing them to the disk directly. This process is very fast.
Of course, such a estimation for the compressed file size is not accurate. To calibrate it, we update the $S_\Delta$ multiple times, as shown in Algo.~\ref{algo:searching}.

\begin{table*}[t]
    \centering
    \resizebox{\textwidth}{!}{
    \begin{tabular}{@{}l|ccccc|ccccc|ccccc@{}}
    \hline
    \multirow{2}{*}{Method} & \multicolumn{5}{c|}{Mip-NeRF 360  - 18.33 MB}     & \multicolumn{5}{c|}{Tank\&Temples - 11   MB}      & \multicolumn{5}{c}{Deep Blending - 8 MB}         \\
                            \cline{2-16}
                            & \makecell{PSNR \\ (dB)$\uparrow$}   & \makecell{SSIM \\ $\uparrow$} &  \makecell{LPIPS \\ $\downarrow$} & \makecell{Size \\ (MB) $\downarrow$ } & \makecell{Time \\ (s) $\downarrow$} &  \makecell{PSNR \\ (dB)$\uparrow$}   & \makecell{SSIM \\ $\uparrow$} &  \makecell{LPIPS \\ $\downarrow$} & \makecell{Size \\ (MB) $\downarrow$ } & \makecell{Time \\ (s) $\downarrow$} &  \makecell{PSNR \\ (dB)$\uparrow$}   & \makecell{SSIM \\ $\uparrow$} &  \makecell{LPIPS \\ $\downarrow$} & \makecell{Size \\ (MB) $\downarrow$ } & \makecell{Time \\ (s) $\downarrow$} \\
                            \hline
    HAC                     & 27.17     & 0.789 & 0.261 & 18.29     & 16627    & \colorbox{red!25}{24.45}     & \colorbox{red!25}{0.854} & \colorbox{red!25}{0.179} & \colorbox{red!25}{10.92}     & 10978     & \colorbox{red!25}{30.27}     & \colorbox{red!25}{0.910} & \colorbox{red!25}{0.254} & 8.29      & 9993     \\
    Our                     & \colorbox{red!25}{27.48}     & \colorbox{red!25}{0.806} & \colorbox{red!25}{0.240} & \colorbox{red!25}{18.17}     &    \colorbox{red!25}{1328}  & 24.04 &	0.840 &	0.200 &	10.93     &     \colorbox{red!25}{1381}     & 30.24 & 0.903  & 0.271  & \colorbox{red!25}{7.92}     &      \colorbox{red!25}{1263}    \\
    \hline
    \end{tabular}
    }
    \caption{\textbf{Performance on size-aware compression}. ``18.33 MB'' refers to the average size budget of the scenes in Mip-NeRF 360. The detailed size target of scenes are listed in supplementary material. The best results are highlighted in \colorbox{red!25}{red} cells.}
    \label{tab:e2e}
    \vspace{-0.8cm}
\end{table*}

\begin{figure*}[t]
    \centering
     \includegraphics[width=\linewidth]{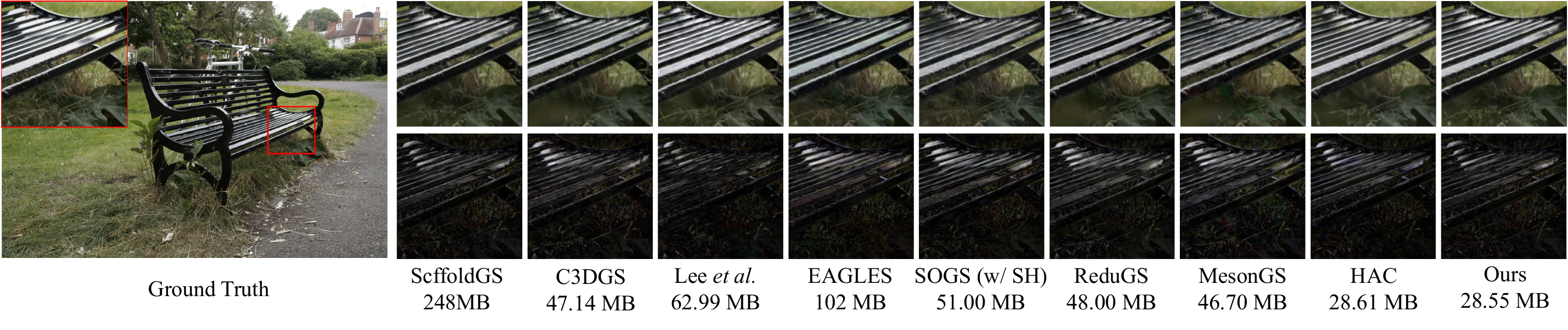}
     \vspace{-0.9cm}
     \caption{\textbf{Qualitative results}. We present the rendering results (rows 1) along with the corresponding error maps (rows 2) from a randomly selected viewpoint of the \textit{bicycle} scene.}
    \label{fig:visual_main}
    \vspace{-0.3cm}
\end{figure*}

\vspace{1mm}\noindent\textbf{Hierarchical Solver.} As there are $(C \times B)^Q$ options for $\rmQ$, if we set $(C, B, Q)$ to $(73, 60, 16)$, a typically setting for compressing ScaffoldGS, then there are $1.8 \times 10^{58}$ options. In such a giant search space, figuring out a suitable bit-width setting is time-consuming. Neither of GPU-based~\cite{fastdog2022} or CPU-based~\cite{pulp} solver can solve the this in minutes. Hence, to accelerate the solving process, we solve this 0-1 ILP problem with two steps. As shown in the left bottom of Fig.~\ref{fig:overview}, at the first step, we search the channel-level bit-width setting $\rmQ_c \in [1, 16]^C$. Then, we calculate the size budget for each channel of attributes based on the $\rmQ_c$. For example, the size budget of channel $i$ can be computed by: $S_c = S_T \frac{\rmQ_{c, i}}{\sum{\rmQ_{c,i}}}$. At the second step, we solve the group-level bit-width setting $\rmQ_g \in [0, 16]^B$ for each channel based on the size limit $S_c$.

\subsection{Piecewise Finetuning}
For compression with finetuning, we directly quantize the coordinates and attributes without applying any transformations to them. To obtain a better compression quality, after the point pruning and coordinate quantization steps, we finetune for multiple epochs respectively to restore the reconstruction quality, as depicted in the right bottom of Fig.~\ref{fig:overview}.

During fine-tuning, we fix coordinates because G-PCC decompression yields unordered points, misaligned with attributes. To align them, we sort both by Morton order after second-stage fine-tuning and build quantization groups on attributes. Changing the coordinates afterward may change the Morton order, which reorders both coordinates and attributes. This reordering breaks the original grouping of attributes used for quantization, making the previously searched bit-widths no longer valid.

\section{Experiments}

\textbf{Datasets.} We conduct experiments on four datasets: 1) Mip-NeRF 360~\citep{barron2022mip}. This dataset contains five outdoor and four indoor scenes. Each scene contains 100 to 300 images. We use the images at 1600×1063. 2) Tank \& Temples~\citep{tandt2017}. This dataset contains the \textit{train} and \textit{truck} scenes. 3) Deep Blending~\citep{hedman2018db}. This dataset contains the \textit{drjohnson} and \textit{playroom} scenes. 4) Synthetic-NeRF~\citep{mildenhall2021nerf}. This is a view synthesis dataset consisting of 8 synthetic scans, with 100 views used for training and 200 views for testing.

\vspace{1mm}\noindent\textbf{Baselines.} We compare our method with the following baselines: 3DGS~\citep{kerbl20233d}, ScaffoldGS~\citep{lu2024scaffold}, C3DGS~\citep{niedermayr2023compressed}, Lee \textit{et al.}~\citep{lee2023compact}, 
LightGaussian~\cite{fan2023lightgaussian},
EAGLES~\citep{girish2023eagles}, SOGS~\citep{morgenstern2023compact}, Compact3D~\citep{navaneet2023compact3d}, ReduGS~\cite{papantonakis2024i3d}, MesonGS~\citep{xie2024mesongs}, HAC~\citep{hac2024}, DVGO~\citep{sun2022direct}, VQRF~\citep{li2023compressing}, and ACRF~\citep{fang2024acrf}. Some of results are derived from HAC~\citep{hac2024}, 3DGS.zip~\cite{3DGSzip2024}, and MesonGS~\citep{xie2024mesongs}, while the visual results are produced by ourselves.

\subsection{Experimental Results}

\textbf{Size-aware Compression on Static Gaussians.} We evaluate end-to-end performance of our method via latency and quality metrics. In Tab.~\ref{tab:e2e}, our method is at most $8\times$ faster than the baseline and achieves better or comparable quality across three datasets. For comparison with the baseline method, we set the number of finetune iterations as 4000. In practical applications, however, fine-tuning for only 500 steps is expected to achieve satisfactory visual quality. We perform a binary search on the hyperparameter $\lambda$ of HAC to find a configuration that meets the size budget. The search stops when the difference between the obtained size and the size budget is within 5\%. The reason why HAC takes so much longer than ours is that, after each hyperparameter adjustment, it requires 10-20 minutes to retrain the mask and context models in order to converge to a specific size. In contrast, our method first searches for the appropriate hyperparameters based on the size, and then only requires a single retraining. Since the hyperparameters are fixed, our method has almost no impact on the file size during retraining.

\vspace{1mm}\noindent\textbf{Size-aware Compression on Dynamic Gaussians. } In the bottom part Tab.~\ref{tab:dynamic}, we applied the SizeGS framework on 4DGS~\cite{yang2024real}. SizeGS can achieves the target size in 51s and improved visual quality after 28s of finetuning. Experiments are conducted on the \textit{flame\_steak} scene of N3DV dataset~\cite{Li_2022_CVPR}. 

\begin{table}
    \resizebox{\linewidth}{!}{%
    \begin{tabular}{@{}l|ccccc@{}}
        \hline
        Method    & PSNR & SSIM & LPIPS & Size     & Finetune Time (s)     \\
        \hline
        MesonGS  & 	25.95 &	0.7706 & 0.2679	& 35.41 MB  & 0     \\
        Our+MesonGS & 26.06 & 0.7743 &	0.2642 & 35.36 MB	& 0	\\
        \hline
        4DGS & 32.06   & - &   - & 5.10 GB   & 0        \\      
        Our+4DGS  & 32.07 &  - & -      & 198.64 MB & 28 \\
        \hline
    \end{tabular}%
    }
    \caption{Quantitative results on 3DGS variants.}
    \label{tab:dynamic}
    \vspace{-1cm}
\end{table}

\begin{figure*}[t]
    \centering
     \includegraphics[width=\linewidth]{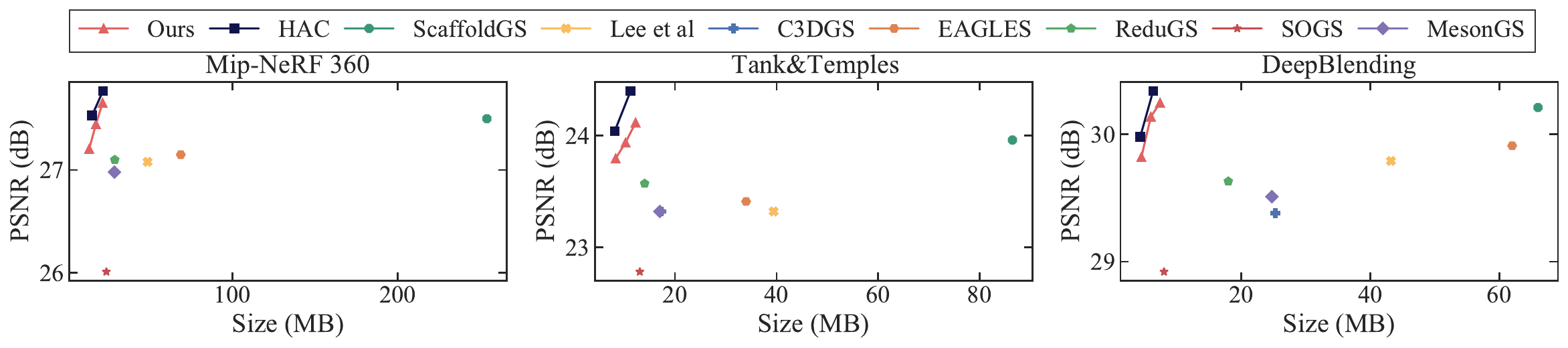}
     \vspace{-0.8cm}
     \caption{\textbf{Rate-distortion curves for quantitative comparison.} Note that our goal is not to improve the marginal performance and defeat the existing compression works. Instead, we aim to design a hyperparameter parameter searching algorithm to compress the 3DGS model into the desired size while maximizing visual quality.}
     \vspace{-0.3cm}
    \label{fig:rd}
\end{figure*}

\vspace{1mm}\noindent\textbf{Performance on Offline Compression.} We implemented our search algorithm on top of the SOTA post-training compression -- MesonGS~\cite{xie2024mesongs}. The results on Mip-NeRF 360 are shown in Tab.~\ref{tab:dynamic}. We can observe that our method enhance the compression quality of MesonGS. For each scene, the size target is set to the compressed file size generated by the official MesonGS configuration.

\begin{table}[]
    \centering
    \begin{tabular}{@{}l|ccccc@{}}
        \hline
        Base model & $\Omega$ & Save  & 01-ILP & Finetune & End2end \\
        \hline
        ScaffoldGS   & 0.11     & 32.78 & 145.22 & 857.71   & 1328.82 \\
        3DGS         & 0.49     & 15.10 & 58.52  & -        & 104.51 \\
        \hline
        \end{tabular}
    \caption{\textbf{Decomposition of time.} The unit of time is second. From the left to right: time to calculate the estimated quality loss $\Omega$, time used to save to storage, time used by ILP solver, time used to finetune the model, and the end-to-end time.}
    \label{tab:time}
    \vspace{-1cm}
\end{table}

\vspace{1mm}\noindent\textbf{Time Breakdown.} Tab.~\ref{tab:time} shows the time breakdown of a round of search. First, the time spent on quality estimation is very short due to parallel quantization. Next, we search $\rmQ$ based on the estimated quality, where size calibration and 0-1 ILP take similar time. After searching for the hyperparameters, if we fine-tune for 6000 iterations, the average time required is approximately 1000s. We set the time limitation for 0-1 ILP to 50s. For ScaffoldGS, we fix the value of $\tau$ to 0.6. For 3DGS, we use the configuration from MesonGS. We can see that it takes 1 minute for ILP to solve the bit-width searching. This is because 3DGS only has 10 channels that are involved in the search, while ScaffoldGS has 73.

\vspace{1mm}\noindent\textbf{Qualitative Evaluation.} In Fig.~\ref{fig:visual_main}, we present the rendering results and the corresponding error maps. From the error maps, it is evident that our method handles chair reflections better than other methods while achieving rendering results that are comparable to ScaffoldGS.

\vspace{1mm}\noindent\textbf{Compared to Online Methods.} The quantitative compression results of different methods are presented in Fig.~\ref{fig:rd}. With enough finetune iterations, our method outperforms most others across all three datasets and achieves performance comparable to leading methods. As implicit neural representations~\cite{zhang2025evos,zhang2025expan,zhang2025sym} serve as an important base model for 3D compression, we also compare with NeRF compression in supplementary material.

\subsection{Ablation Study}
Unless stated otherwise, all experiments use the \textit{bicycle} scene from Mip-NeRF 360 and ScaffoldGS as the base model.

\begin{table}[t]
    \centering
    \resizebox{\linewidth}{!}{
    \begin{tabular}{@{}l|c|ccc@{}}
        \hline
        Method                 & Budget (B)           & Searched (B) & $\Delta$ size (B) & Information loss \\
        \hline
        GA & \multirow{3}{*}{$3\times 10^7$} & 21,833,128        & 8,166,872      & 42,821,038         \\
        Vanilla ILP           &                           & 28,934,805        & 1,065,195      & 1,258,394          \\
        0-1 ILP (Our)               &                           & 29,831,203        & 168,797        & 11,826      \\
        \hline
    \end{tabular}
    }
    \caption{\textbf{Superiority of 0-1 ILP.} ``GA'': Genetic Algorithm. With up to 16 bit-width choices, the Vanilla ILP and GA that are widely adopted in model quantization methods are unable to quickly search for suitable mixed-precision settings.}
    \label{tab:ilp_aba}
    \vspace{-0.5cm}
\end{table}

\vspace{1mm}\noindent\textbf{0-1 ILP Superiority in Searching Bit-widths.} In solving the optimal bit-width setting for different attribute channels, we also demonstrate the superiority of the 0-1 ILP. As shown in Tab.~\ref{tab:ilp_aba}, we experiment with widely-used General ILP~\citep{yao2021hawq} and genetic algorithms~\citep{guo2020single,tang2024retraining}, both of which proved inferior. The 0-1 ILP fully utilizes the size budget while minimizing information loss. General ILP involves variables ranging from 1 to 16. In contrast, 0-1 ILP’s binary values offer finer control, easier integration of constraints, and more efficient solution techniques. Genetic algorithms, though suited for black-box problems, handle constraints less efficiently, making them unsuitable for our problem.

\begin{figure}[t!]
    \centering
    \begin{subfigure}{0.9\linewidth}
        \includegraphics[width=0.99\linewidth]{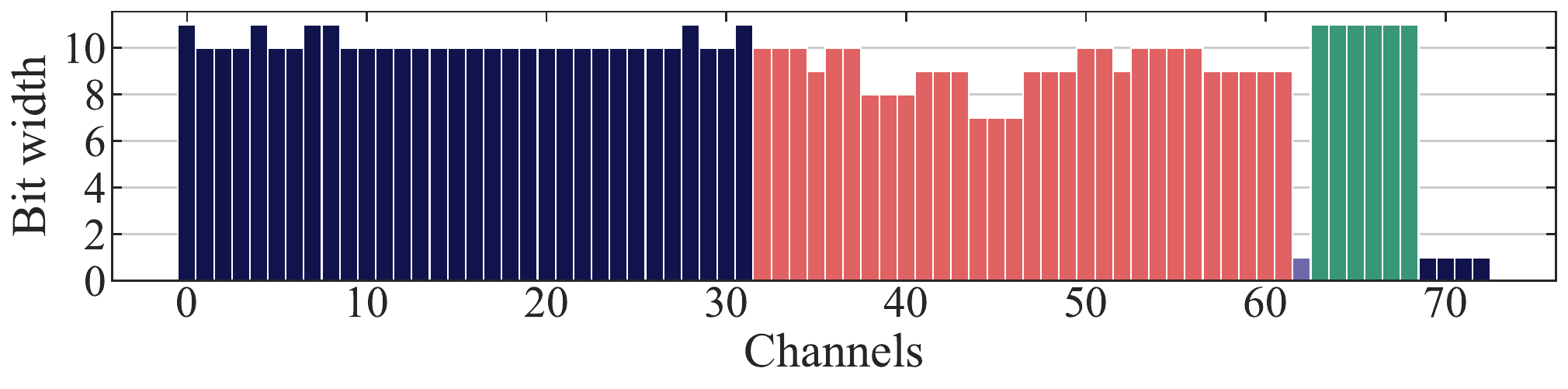}
        \vspace{-0.2cm}
      \caption{Budget: 28 MB, Scene: \textit{bicycle}.}
      \label{fig:bw28}
    \end{subfigure}
    \hfill
    \begin{subfigure}{0.9\linewidth}
        \includegraphics[width=0.99\linewidth]{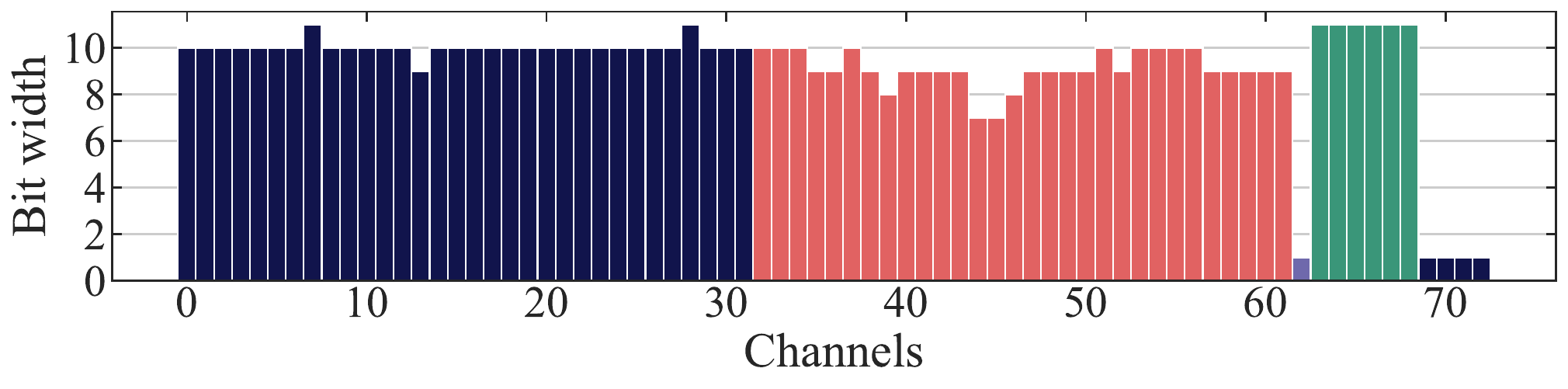}
        \vspace{-0.2cm}
      \caption{Budget: 12 MB, Scene: \textit{bicycle}.}
      \label{fig:bw12}
    \end{subfigure}
    \vspace{-0.3cm}
    \caption{\textbf{Bit widths of each channel.} We use different colors to represent different kinds of attributes. From left to right are: \textcolor{MyBlue}{features $\rvf$}, \textcolor{MyRed}{offsets $\rmO$}, \textcolor{MyPurple}{opacity $o$}, \textcolor{MyGreen}{scaling $\rvl$}, and \textcolor{MyBlue}{rotation $\rvr$}.}
    \label{fig:bit_widths}
    \vspace{-0.2cm}
\end{figure}

\vspace{1mm}\noindent\textbf{Bit-widths.} In Fig.~\ref{fig:bit_widths}, we show the bit-widths for different attribute channels, where different colors represent different attributes. Here, we assume that the quantization groups that belong to the same channel share the same bit-width setting.
From left to right in order: features $\rvf$, offsets $\rmO$, opacity $o$, scaling $\rvl$, and rotation $\rvr$. We can see that under different size budgets, the choices of bit-widths are generally the same.

\begin{figure}[t]
    \centering
    \begin{subfigure}{0.48\linewidth}
        \includegraphics[width=0.99\linewidth]{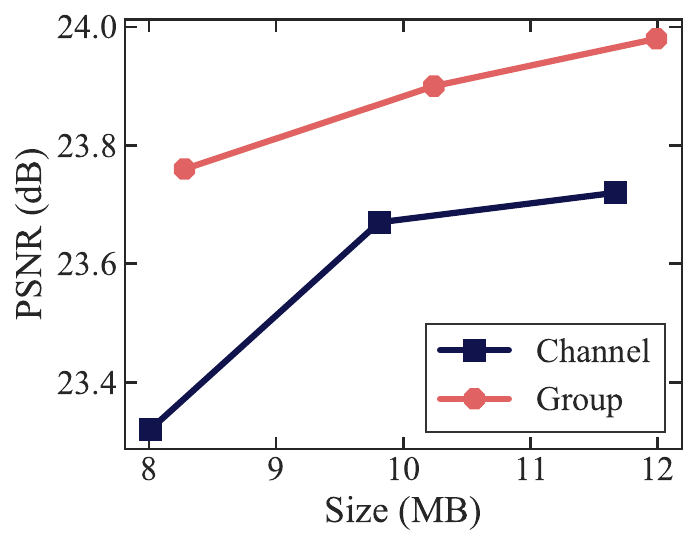}
      \caption{Group vs. Channel}
      \label{fig:diff_comp}
    \end{subfigure}
    \hfill
    \begin{subfigure}{0.48\linewidth}
        \includegraphics[width=0.99\linewidth]{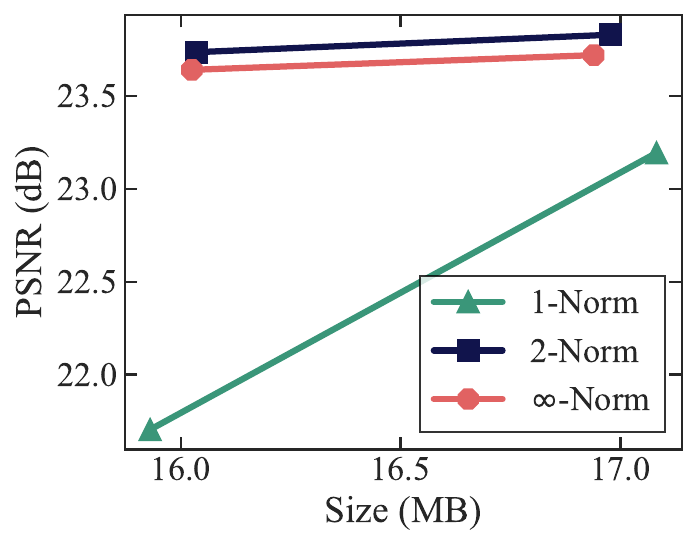}
      \caption{Different norms for $\Omega$.}
      \label{fig:aba_loss}
    \end{subfigure}
    \vspace{-0.2cm}
    \caption{\textbf{Ablation studies.} (a) Results were conducted with sufficient finetuning, confirming that mixed bit-width settings enhances the upper bound of compression quality. (b) 2- and $\infty$-norms significantly outperform 1-norm.}
  \vspace{-0.15cm}
\end{figure}

\vspace{1mm}\noindent\textbf{Effectiveness of Mixed Bit-width Setting.} An alternative is letting the groups that belong to the same channel share the same bit-width, like the first step of hierarchical solver in Sec.~\ref{sec:ilp}. To verify the necessity of group-wise bit-width setting, we compare the effectiveness of channel-wise and group-wise mixed bit-widths setting in Fig.~\ref{fig:diff_comp}, both with sufficient finetuning. Results show that finer granularity yields better performance.

\begin{table}[t]
    \resizebox{\linewidth}{!}{%
    \begin{tabular}{@{}ccc|cccc@{}}
    \hline
    Prune & Voxel & Quant & PNSR (dB) $\uparrow$ & SSIM $\uparrow$  & LPIPS  $\downarrow$& Size (MB) $\downarrow$ \\
    \hline
    0     & 0     & 6000  & 24.95 & 0.7331 & 0.2709 & 24.20     \\
    1000  & 0     & 5000  & 24.98 & 0.7339 & 0.2706 & 24.19     \\
    0     & 1000  & 5000  & 24.99 & 0.734  & 0.2707 & 24.12     \\
    1000  & 1000  & 4000  & 25.19 & 0.7451 & 0.2619 & 24.25     \\
    \hline
    \end{tabular}
    }
    \caption{Efficiency of piecewise finetuning. ``Prune'', ``Voxel'', and ``Quant'' refer to the finetune iterations after the three part. We fix the total iterations as 6000.}
    \label{tab:ft}
    \vspace{-0.8cm}
\end{table}

\vspace{1mm}\noindent\textbf{Effectiveness of Piecewise Finetune.} In Tab.~\ref{tab:ft}, we evaluate performance of finetuned 3DGS under four iteration settings. We set the target size as 24 MB. Piecewise fine-tuning helps improve the upper bound of compression quality.

\vspace{1mm}\noindent\textbf{Estimated Quality Loss $\Omega$.} There are many ways to calculate the distance between the original attributes and the restored attributes. To investigate which metric is better, in Fig.~\ref{fig:aba_loss}, we show the impact of different metrics on the information loss of the final rendering results. We present the PNSR-Size curves under three metrics, including 1-norm, 2-norm, and $\infty$-norm. It can be seen that the performance of 2-norm and $\infty$-norm are nearly the same, both of which outperform 1-norm by a significant margin.

\begin{table}[t]
    \centering
    \resizebox{\linewidth}{!}{
    \begin{tabular}{@{}l|c|ccc|c@{}}
    \hline
    $K$  & Budget (MB)  & PSNR  (dB)$\uparrow$  & SSIM  $\uparrow$ &  LPIPS $\downarrow$ & Searched Size (MB) $\downarrow$ \\
    \hline
    40        & \multirow{3}{*}{30} & 25.13  & 0.7410 & 0.2684 & 29.85          \\
    30        &                        & 25.15  & 0.7411 & 0.2685 & 29.91          \\
    50        &                        & 25.14  & 0.7413 & 0.2686 & 29.86          \\
    \hline
    40        & \multirow{3}{*}{20} & 25.07   & 0.7353 & 0.2752  & 19.83          \\
    30        &                        & 25.07   & 0.7357 & 0.2757  & 19.85          \\
    50        &                        & 25.12   & 0.7368 & 0.2743  & 19.92          \\
    \hline
    \end{tabular}
    }
    \caption{\textbf{Robustness Evaluation.} The bit-width setting can adapt to different values of the number of blocks $K$, ensuring the visual quality within a given size is not affected by $K$.}
    \label{tab:robust}
    \vspace{-0.8cm}
\end{table}

\vspace{1mm}\noindent\textbf{Robustness Evaluation}. We evaluated the size and corresponding performance of the searching algorithm under different numbers of blocks. As shown in Tab.~\ref{tab:robust}, for varying numbers of blocks and different target sizes, our method consistently finds appropriate bit-width settings, ensuring that the final file size is close to the target while maintaining optimal visual quality. Regardless of the block number setting, the final file size and performance are similar, indicating that our method is robust to the number of blocks.

\section{Related Work}

\subsection{3D Gaussian Splatting and Its Compression}
3D Gaussian Splatting (3DGS)~\cite{kerbl20233d} achieves excellent 3D reconstruction but suffers from large model size. Early efforts~\cite{niedermayr2023compressed,lee2023compact,girish2023eagles,morgenstern2023compact,navaneet2023compact3d,papantonakis2024i3d,fan2023lightgaussian,xie2024mesongs,wang2024rdo,liu2024compgs,wu2024implicitgs,tang2025neuralgs} focused on compressing the original 3DGS. Later, works~\cite{hac2024,lee2025opt,liu2024hemgs} targeted more efficient 3DGS variants~\cite{fang2024mini,lu2024scaffold,ali2024trimming,ren2024octree,zhan2025catdgs}, especially ScaffoldGS~\cite{lu2024scaffold}, which groups anchors into voxels and uses per-voxel features to predict Gaussians. HAC~\cite{hac2024,hac++2025} leverages 3D coordinates to guide quantization and entropy coding. ContextGS~\cite{wang2024contextgs} encodes anchors hierarchically. HEMGS~\cite{liu2024hemgs} uses a pretrained PointNet++\cite{qi2017pointnet} for better context modeling. FCGS\cite{fcgs2024} compresses 3DGS via one-shot inference but supports only fixed-size output and lacks adaptability to other 3DGS forms. Some works~\cite{pcgs2025,disario2025gode,shi2024lapisgs} address bitrate fluctuation via layered coding, which is orthogonal to our approach.

Unlike the previous works, our work propose to configure the hyperparameters of 3DGS from the perspective of combinatorial optimization and design many techniques to accelerate the process.

\subsection{Mixed Precision Quantization}
Our work compress the 3DGS to desired size by searching for two hyperparameters: one is $\tau$, and the other is the bit-width setting $\rmQ$. The optimization for the bit-width setting is closely related to mixed precision quantization for deep learning models.

Mixed Precision quantization (MPQ) is a widely-used technique to improve the trade-off between the accuracy and efficiency of neural networks~\cite{wang2018haq,dong2019hawq,dong2019hawqv2,yao2021hawq,tang2022mixed}. The challenge with this approach is to find the right mixed-precision setting for the different layers of neural networks. A brute force approach is not feasible since the search space is exponentially large in the number of layers. HAQ~\cite{wang2018haq} employed reinforcement learning to search this space. However, this RL-based solution requires tremendous computational resources. HAWQ~\cite{dong2019hawq,dong2019hawqv2,yao2021hawq} proposed to assign each layer a sensitivity score with the Hessian spectrum and then use an ILP solver to generate mixed-precision settings with various constraints (such as model size and latency). Though CA-NeRF~\cite{liu2024content} uses the MPQ scheme, their method cannot be applied to 3DGS. 

Unlike the HAWQ3~\cite{yao2021hawq} that uses the ILP, we use binary ILP to search a better results for the bit-width settings of 3DGS compression. Though some works~\cite{hac2024,liu2024compgs} employed MPQ, their required retraining for configuring the quantization settings.

\section{Conclusion}
In this paper, we present \emph{SizeGS}, a method for automatically selecting hyperparameters to compress 3D Gaussians to a target file size while maximizing visual quality. We formulate this problem as a mixed integer non-linear programming model and decouples this problem into two steps: discrete sampling of reserve ratio $\tau$ and ILP for bit-width settings $\rmQ$. To accelerate the ILP process, we use quantization loss to replace the quality metric and write a CUDA kernel to parallelize the quantization. We also design a size estimator to help search out a more accurate hyperparameter set. Experiments show SizeGS effectively controls file size while preserving quality.

\clearpage
\begin{acks}
  We sincerely thank the anonymous reviewers from ICLR, CVPR, and ACM MM for their valuable feedback and suggestions. We also thank our lab mates for their help in improving the manuscript. This work is supported in part by National Key Research and Development Project of China (Grant No. 2023YFF0905502), National Natural Science Foundation of China (Grant No. 92467204 and 62472249), and Shenzhen Science and Technology Program (Grant No. JCYJ20220818101014030 and KJZD20240903102300001). 
\end{acks}

%

\bibliographystyle{ACM-Reference-Format}
\balance
\bibliography{sample-base}

\end{document}